\documentclass[conference]{IEEEtran}
\ifCLASSINFOpdf
   \usepackage[pdftex]{graphicx}
  \DeclareGraphicsExtensions{.pdf,.jpeg,.png,}
\else
  \usepackage[dvips]{graphicx}
  \DeclareGraphicsExtensions{.eps}
\fi
\usepackage{xcolor}
\usepackage{soul}

\usepackage{amsmath,amssymb}
\usepackage[caption=false,font=footnotesize]{subfig}

\usepackage{balance}

\begin{document}
\title{Towards Deep Physical Reservoir Computing \\ Through Automatic Task Decomposition And Mapping}

%\author{\IEEEauthorblockN{Matthias Freiberger\\ and Joni Dambre}

%\IEEEauthorblockA{

%IDLab, \\
%Ghent University - imec,\\
%Technologiepark 126,\\
%9052 Gent, Belgium}
%\and
%\IEEEauthorblockN{Peter Bienstman}
%\IEEEauthorblockA{
%Photonics Research Group, \\

%Ghent University - imec,\\
%Technologiepark 126,\\
%9052 Gent, Belgium}
%}


\author{\IEEEauthorblockN{Matthias Freiberger\IEEEauthorrefmark{1},
%Andrew Katumba\IEEEauthorrefmark{2},
Peter Bienstman\IEEEauthorrefmark{2} and
Joni Dambre \IEEEauthorrefmark{1}}
\IEEEauthorblockA{\IEEEauthorrefmark{1} IDLab-AIRO \\ Ghent University - imec \\ Technologiepark Zwijnaarde 126, 9052 Gent, Belgium}
\IEEEauthorblockA{\IEEEauthorrefmark{2} Photonics Research Group  \\Ghent University - imec \\ Technologiepark Zwijnaarde 126, 9052 Gent, Belgium}}

%Telephone: (800) 555--1212, Fax: (888) 555--1212}


\maketitle

\begin{abstract}
Photonic reservoir computing is a promising candidate for low-energy computing at high bandwidths. Despite recent successes, there are bounds to what one can achieve simply by making photonic reservoirs larger. Therefore, a switch from single-reservoir computing to multi-reservoir and even deep physical reservoir computing is desirable. Given that backpropagation can not be used directly to train multi-reservoir systems in our targeted setting, we propose an alternative approach that still uses its power to derive intermediate targets. In this work we report our findings on a conducted experiment to evaluate the general feasibility of our approach by training a network of 3 Echo State Networks to perform the well-known NARMA-10 task using targets derived through backpropagation. Our results indicate that our proposed method is well-suited to train multi-reservoir systems in a efficient way.  
\end{abstract}
% \fi
\IEEEpeerreviewmaketitle

\section{Introduction}
Reservoir computing was proposed in the early 2000s to circumvent difficulties when training recurrent neural networks \cite{Maas2002,Jaeger2004}. More recently, it has gained increased interest in the unconventional computing community as it proves to be an excellent framework to train physical nonlinear dynamic systems to perform meaningful computations \cite{Buerger2013,Demis2015,hermans2015photonic}. With photonic reservoir computing, which uses light processing substrates as reservoirs, competitive results have been achieved on time series prediction, voice recognition, non-linear distortion compensation and telecommunication tasks  \cite{Soriano13,smerieri2012towards,Vandoorne2014, katumba2019neuromorphic}. In combination with currently researched technologies on non-volatile optical weights \cite{Abel2013Astrong}, this technology is a promising candidate to  pave the way to efficient low-energy computing at large bandwidths ($> 10$ Gbps). 

Nevertheless it is well known in the reservoir computing community that there are bounds to what one can achieve simply by making a reservoir larger and using more observed states in the readout. This is mostly due to the fact that the states are usually highly correlated due to the interactions inside the reservoir. This makes it difficult to exploit the residual information that is added by additional states as the state space grows larger.  For simulated reservoirs, such as infinite precision echo state networks, it would require an increase of the amount of training data. In physical reservoirs, noise and measurement inaccuracies effectively constrain the exploitation of state information and performance tends to saturate even faster as a function of reservoir size. In addition, fabricating larger integrated (photonic) reservoirs increases the technological challenge, e.g., due to routing problems and high optical losses as well as cost and yield-related constraints on the chip size.  In summary, in order to move to more complex tasks, simply making reservoirs larger is not enough. Instead, there is a need for approaches to design multi-reservoir or even deep reservoir systems. Gallicchio et al. \cite{gallicchio2017deep} propose to scale reservoir computing by building a large reservoir out of a set of smaller reservoirs, where each reservoir injects its output into the next reservoir without training intermediate connections. The states of all reservoirs are then fed into the readout. While it performs well, this approach is challenging to implement with integrated photonic reservoirs due to its high wiring effort, which subjects it to almost the same routing issues and constraints as for large monolithic reservoirs.
\begin{figure}[h!]
\centering
 \includegraphics[width=2.3in]{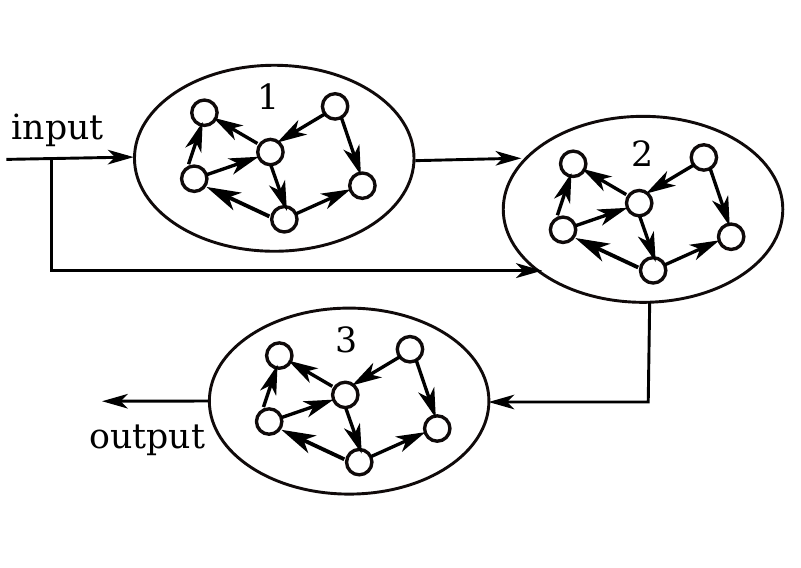}
\label{fig:ensemble} \\ \vspace{-0.5cm}
\caption{System Architecture of 3 connected Echo State Networks for which intermediate desired signals have been derived using backpropagation.}  
\label{fig:architectures}
\end{figure}

An alternative approach is to individually train many smaller, interconnected reservoirs, which collaborate in a network to solve the task at hand. Two possible multi-reservoir architectures, based on standard ensembling approaches from machine learning, have been evaluated in \cite{freiberger2018improving}. Yet, we expect better performance can be achieved by a tailored network of reservoir modules for each task. Unfortunately, there is no general way to derive desired signals necessary to train the intermediate outputs of each reservoir in such an architecture. Using backpropagation to train the readouts in a network of physical reservoirs in simulation may appear as an obvious solution. However, the differences between individual integrated coherent photonic reservoirs due to process variations are so large that readout weights trained in simulation are not transferable. As a consequence, each reservoir or network of reservoirs must be trained individually on the physical hardware. A mass-market maturity of integrated reservoirs is only achievable if large numbers of devices can be trained in an acceptable time frame. 
 
 To a degree, designing multi-reservoir systems poses a solution to this problem. A given class of tasks can be decomposed by hand into a combination of smaller subtasks, which can be mapped onto a corresponding network of reservoirs that are trained on these subtasks. Unfortunately, this approach has two drawbacks: (i) it can not be guaranteed that a suitable decomposition exists for each given task and (ii) there is no general approach to finding an engineered decomposition. Given that backpropagation can not be used directly to train multi-reservoir systems in our targeted setting, we propose an alternative approach that still uses its power to derive intermediate targets. Our approach is to define a seemingly suitable architecture of reservoirs and attempt to train this architecture in simulation using backpropagation. The resulting trained task decomposition works well for the specific reservoirs that were simulated. We then assume that reservoirs from the same class (i.e., with identical design parameters) can approximate functions of the input history with similar accuracy. Under this assumption, the readout signals from the simulated reservoirs can be used as training targets for physical reservoirs in the same network configuration.

In general, there is no guarantee that an appropriate task decomposition exists for a given network and task. To evaluate the feasibility of our approach, we therefore provide a proof-of-concept with a benchmark task for which we were able to find a good engineered task decomposition, i.e., one for which the decomposed system has similar performance as a monolithic reservoir. We decompose the input-output relation analytically and map the resulting individual components to the connected reservoirs shown in Figure \ref{fig:architectures}. The resulting trained system is then used as a baseline to evaluate our approach, using simulated echo state networks (ESN) \cite{Jaeger2004} as reservoirs. We find that backpropagation trains desired signals that approximate our hand-designed signals where similarity increases with the proximity to the output layer. Finally, we successfully use the intermediate signals found as targets for a classic reservoir training approach. 
%:  we train a third set of networks with the same architecture and design parameters using ridge regression. The outputs of the intermediate stages of the system trained with backpropagation are now used as desired signals at these stages. We find that systems trained in such a way exhibit performance comparable to the backpropagated systems from which it's intermediate desired signals were derived. In a nutshell, our findings indicate that backpropagation may be suited to derive intermediate targets for physical multi-reservoir architectures, which are then to be trained with classic training algorithms.

\section{Methodology}
{\bf Task} In this work we use the Nonlinear Autoregressive Moving Average system identification task of order 10 (NARMA-10) \cite{atiya2000new} as a benchmark for our new training approach. Its desired signal is defined as follows:
\begin{equation}
\label{eq:desired-original}
\begin{split}
y[n+1] = &  \thinspace 0.05 \thinspace  y[n] \thinspace \sum_{k=0}^9 y[n-k] +  \\
                 &  \thinspace 0.3 \thinspace  y[n] +  1.5  \thinspace u[n - 9] \thinspace u[n] + 0.1,
\end{split}
\end{equation}
where $y[n]$ denotes the desired output signal and $u[n]$ denotes a random input sequence uniformly distributed in $[0,0.5]$. For $n < 0$, $u[n]$ and $y[n]$ are by definition $0$. 

We decomposed this task into three subtasks to be mapped on the reservoirs in the multi-reservoir network of Figure \ref{fig:architectures}. As there are delayed input components in the task, a first subtask (Module 1) is a delay line with the desired signal
\begin{equation}
y_{1}[n] = u[n - 9].
\end{equation}
Module 2 multiplies the delayed version of the input signal with its original version:
 \begin{equation}
y_{2}[n] = u[n] \thinspace  \hat{y_{1}}[n], 
\end{equation}
where $\hat{y_{1}}[n]$ is the output module 1, which only approximates its desired output. Finally, module 3 converts the output of module 2 into the final NARMA-10 output: 
\begin{equation}
\label{eq:desired3}
\begin{split}
y_3[n+1] =  & \thinspace 0.05 \thinspace y_{3}[n] \sum_{k=0}^{9} y_{3}[n-k] + \\
                      & \thinspace 0.3 \thinspace y_{3}[n] + 1.5 \thinspace \hat{y_2}[n] + 0.1,  
\end{split}
\end{equation}
where $\hat{y_2}[n]$ is the true output of module 2. If the approximations made by the three modules are perfect, this decomposition exactly reproduces the original task. 

For all experiments, we used the normalized mean squared error (NMSE) between each architecture's output signal and the desired signal as generated by Equation \ref{eq:desired-original}:
\begin{equation}
e_{\text{NMSE}} = \frac{1}{N} \sum_{n=1}^{N} \frac{(\hat{y}[n] - y[n])^2}{\sigma_y^2},
\end{equation}
where $\hat{y}[n]$ is the prediction of the trained network, and $\sigma_y^2$ is the variance of the desired signal $y[n]$.

{\bf Architectures} As modules in the multi-reservoir architecture of Figure \ref{fig:architectures}, we used ESNs of 100 nodes. Because the aim was to use backpropagation on the architecture, we used Exponential Linear Units (ELUs) \cite{clevert2015fast} as nonlinearities to ensure a smooth propagation of gradients. 

To verify whether our engineered solution is competitive, we compared to a monolithic reservoir with similar resources, i.e., an ESN with 300 nodes. While the main point of this work is not to outperform previous training approaches and networks on NARMA-10, but rather finding an efficient way to train multi-reservoir systems, we consider establishing a simple baseline as a point of reference mandatory. We use a hyperbolic tangent nonlinearity for the single-ESN system since this nonlinearity exhibited the best performance for single-reservoir systems in preliminary experiments. 

We trained our multi-reservoir system in 3 different ways. Firstly, we evaluated the engineered task decomposition by training the three ESNs in the system to fit the three desired signals respectively. The ESNs were trained incrementally: each ESN takes input from previously trained ESNs and/or the input signal to solve its assigned subtask. This way, each ESN is trained to be maximally robust to approximation inaccuracies of its predecessor(s).  

Secondly, after having shown that the NARMA-10 task can be solved by decomposing it by hand and mapping it to a suitable reservoir architecture, we trained the multi-reservoir architecture from the previous step (using different, randomly initialized ESNs) as a whole using backpropagation to automatically find a good task decomposition. 
Thirdly, we evaluate the transferability of the learned decomposition. Here, the best performing network obtained in the previous step was used to generate target signals for networks of different, randomly initialised ESNs from the same class. 

{\bf Training and parameter tuning} To train our networks, we generated train, validation and test set time series of 100000 samples each. The regularization strength of our ridge regression reservoir readouts was  determined by performing grid search in combination with 5-fold cross-validation on the train set. For both the monolithic ESN as well as mapping the engineered task decomposition onto the multi-reservoir architecture, the validation set was used to find the hyperparameters that are intrinsic to the reservoir: spectral radius, input scaling, and bias scaling, as well as the sparsity of the reservoir input and connection matrices. These were tuned using hyperopt \cite{bergstra2015hyperopt}. The resulting hyperparameter values were then transferred to initialise the architectures used to test the automatic task decomposition.

When backpropagating through multi-ESN architectures, the necessary hyperparameters for that algorithm, i.e. learning rate, batch size, weight decay and amount of gradient noise were tuned on the validation set. The performance of all approaches was compared on the test set. 

 We have trained our network using backpropagation through time with a batchsize of 60 samples for 120 epochs. We used Adam \cite{kingma2014adam}, starting with a learning rate of $0.0005$. After 60 epochs we divided the learning rate by 2. We applied a weight decay $\lambda = 0.0001$. We have used standard deep learning techniques to improve training convergence for our networks: gradient clipping  \cite{pascanu2013difficulty}, batch normalisation \cite{ ioffe2015batch}, and gradient noise \cite{neelakantan2015adding}. Mind that we only use techniques which can be applied in simulated networks of integrated photonic reservoirs as well. While we acknowledge that our method stands or falls with the ability to successfully backpropagate through such networks in simulation, this appears feasible since Hermans et al. \cite{hermans2015trainable} have already demonstrated backpropagation through a similar systems.

 {\bf Analysis} After achieving a satisfactory task performance on our architecture, we compared the trained intermediate output signals of the individual reservoirs in our architecture with the desired signal we derived by hand. This was done as a first investigation into the task decomposition that was found with backpropagation. 

\begin{figure*}[!t]
\centering
\subfloat[ESN 1 correlation: engin. vs. trained targets]{\includegraphics[width=2.3in]{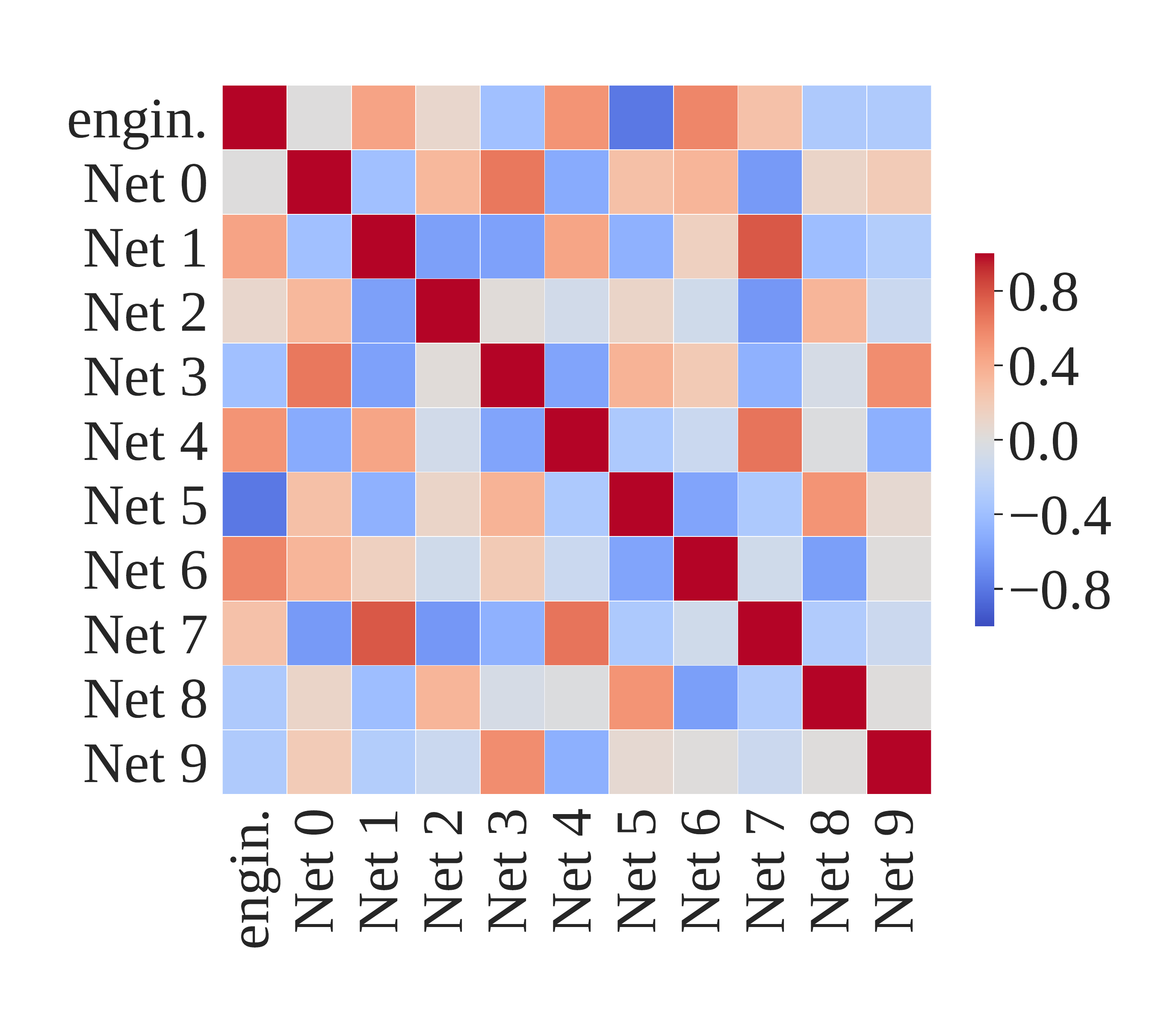}%
\label{fig:esn_1_corr}}
\hspace{2.1cm}
\subfloat[ESN 2 correlation: engin. vs. trained targets]{\includegraphics[width=2.3in]{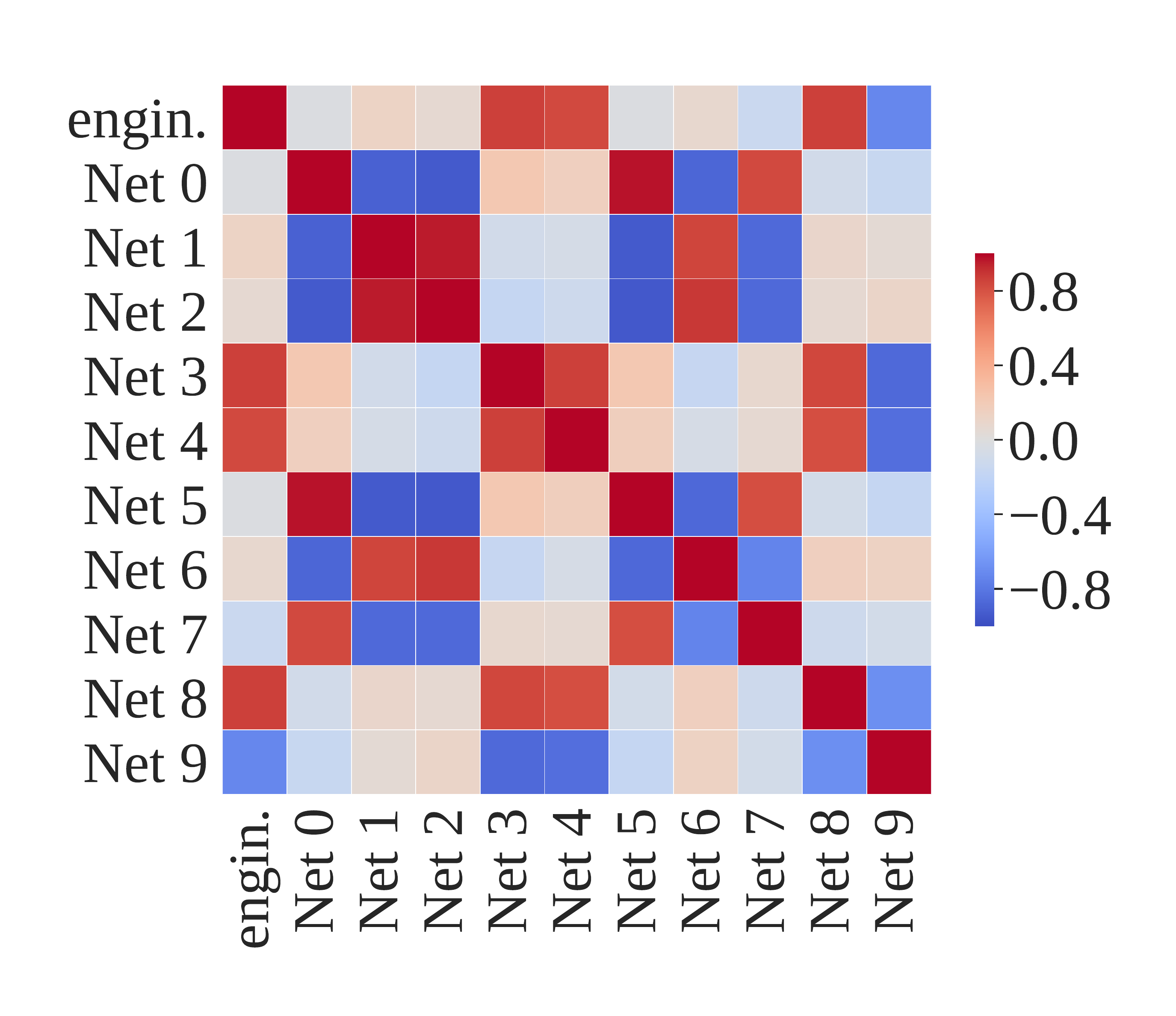}%
\label{fig:esn_2_corr}} \\
\vspace{-0.4cm}
\subfloat[ESN 1: engin. vs. trained signals]{\includegraphics[width=3.5in]{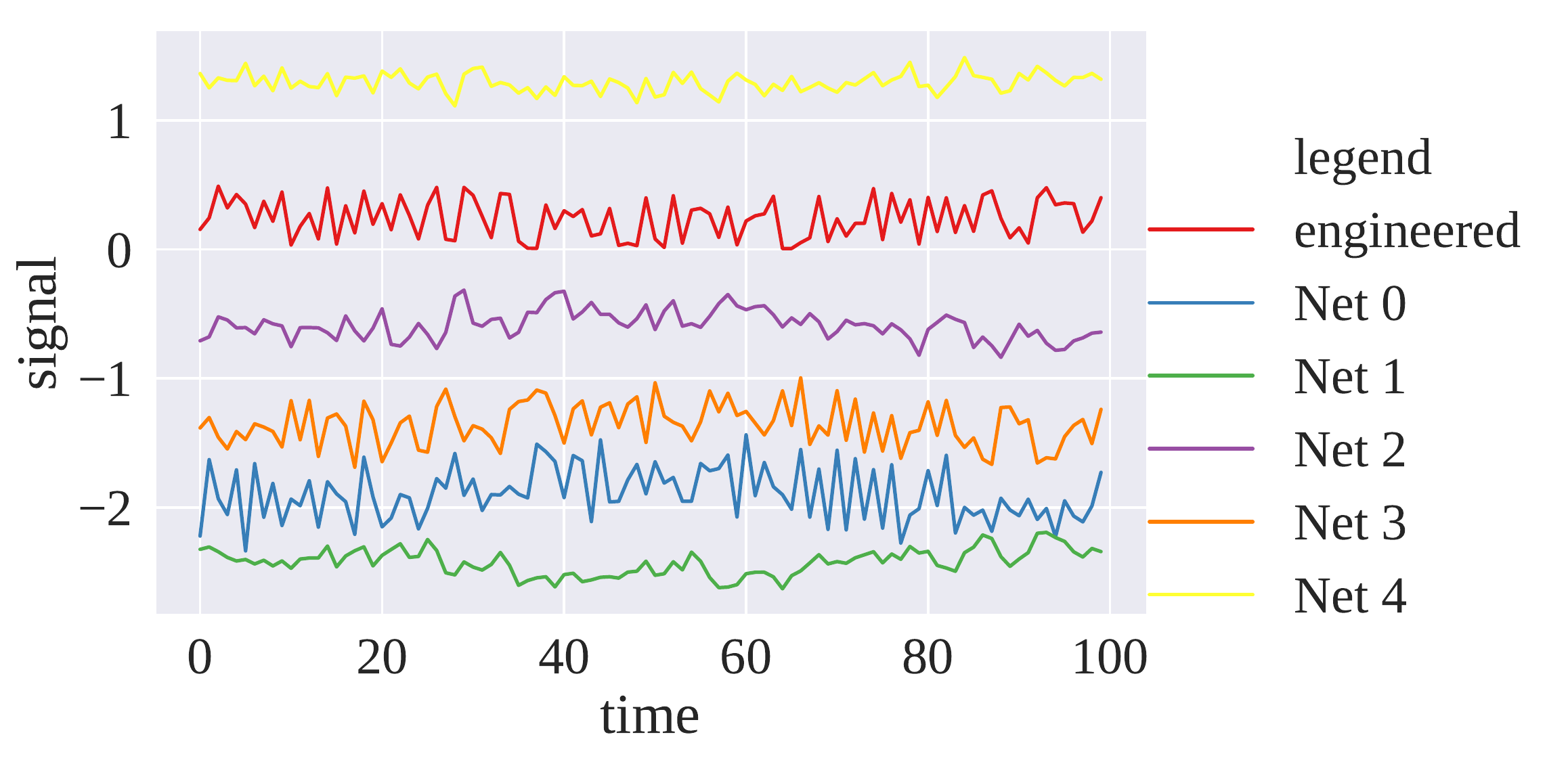}%
\label{fig:esn_1_output}}
\subfloat[ESN 2: engin. vs. trained signals]{\includegraphics[width=3.5in]{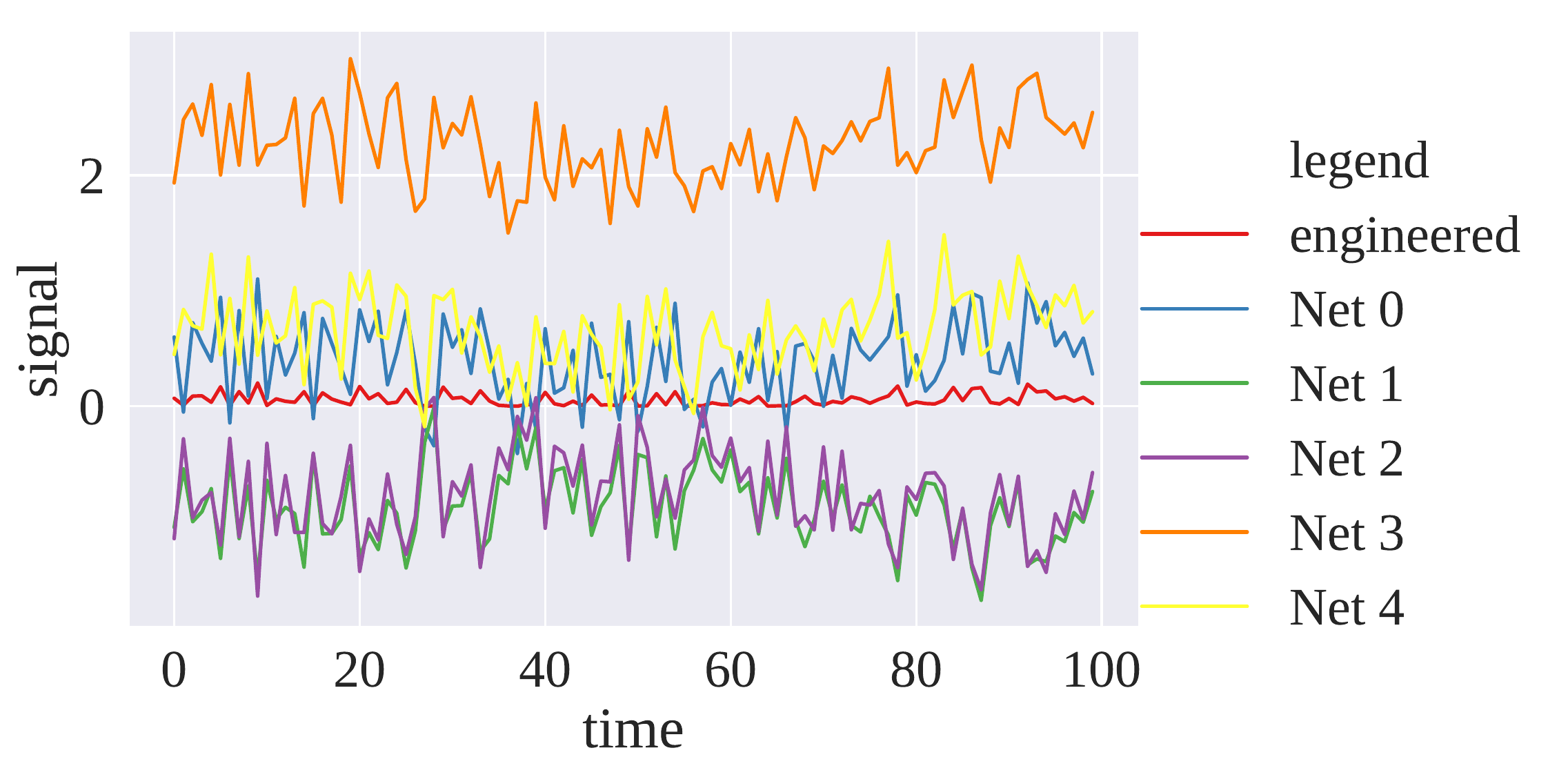}%
\label{fig:esn_2_output}}

\caption{Figure \ref{fig:esn_1_corr} and \ref{fig:esn_2_corr} show the correlation matrix between engineered desired signal and trained desired signals using backprop for ESN 1 and ESN 2 respectively.  Figure \ref{fig:esn_1_output} and Figure \ref{fig:esn_1_output} show the trained target signals of 5 trained multi-ESN systems for ESN 1 and ESN 2 respectively.}%In the signal plots all signals have been normalized to have a standard deviation of 0.5 for better comparison to each other}
\label{fig:stacking}
\end{figure*}

\section{Results}
We started out by training 10 monolithic ESNs as well as 10 engineered multi-ESN systems, each comprising three 100 node ESNs with different random weight initializations. Our monolithic ESN system achieves an average NMSE of $0.035$ with a standard deviation of $0.016$ on the test set, which is close to the results reported in \cite{jaeger2003adaptive}. Note that, while the ESNs in \cite{jaeger2003adaptive} are smaller, their reservoir state matrix is concatenated with a squared version of itself prior to training/evaluation in order to enhance state richness and utilize an additional output nonlinearity. We omit these steps in favor of simplicity and to ensure that our results allow us to judge the feasibility of our approach for future hardware implementations, where such operations would need to be omitted or implemented by a circuit. 

As intended, the engineered multi-ESN system slightly outperforms the single ESN baseline with an average test NMSE of $0.027$ at a standard deviation of $0.015$. After establishing the quality of our engineered task decomposition, we trained 10 new multi-reservoir systems using the same architecture and ESN hyperparameters of our engineered system, but with new, randomly initialized ESNs. The best of those models achieves an NMSE of $0.039$. However, we find that the performance of our models varies a lot, due to large variations in the convergence of the backpropagation. These convergence issues result in an average NMSE of $0.076$ with a standard deviation of $0.024$. This indicates that there is still a lot of room for improvement here, for example by identifying more suitable learning rate adaption schemes and even more rigorous tuning of all training hyperparameters in general. This will be a topic for future work. What is commonly done when learning convergence varies a lot across training runs is simply to select the best solution across multiple runs (which is what we will do here). We found that the achieved performance of the best solution found (referred to as Net 0 in all plots of this section) is sufficient to investigate the trained system and deliver a proof of concept for our proposed training scheme. 

To gain more insight into how the intermediate signals found by backpropagation relate to the engineered desired signals, we computed the correlation matrix between the engineered desired signals and the intermediate signals for ESNs 1 and 2 for all 10 training runs. Figure \ref{fig:esn_1_corr} and \ref{fig:esn_2_corr} show the results. For ESN 1, the signals are at best weakly correlated: only signals 4, 5 and 6 show some correlation (positive or negative) to the engineered signal. For ESN 2, things look different: here we can see strong correlations or anti-correlations to the engineered target signal for 4 out of the ten training runs. In addition, overall, there seem to be two clusters. The signals for runs 3, 4, 8 and 9 are correlated to the engineered target signal, and the signals for runs 0, 1, 2, 5, 6 and 7  are strongly correlated/anticorrelated with each other but show almost no correlation with the engineered target signal. This suggests that for these cases, a different task decomposition with similar performance was found. 

We further investigated this by plotting 100 timesteps of the trained signals for the first 5 training runs as seen for ESN 1 and ESN 2 in Figure \ref{fig:esn_1_output} and \ref{fig:esn_2_output}, respectively. For the output of ESN 1, the plots confirm what was seen in the correlation plots, i.e. that there are no strong similarities between these signals. For ESN 2 we see that indeed there seem to be 2 different decompositions, where the main difference between the decompositions appears to be delay. Consider the difference between Net 0 and Net 4, who belong to different correlation clusters. The two signals are of similar shape, but show a clear time shift between them. This suggests that the main difference between the two decompositions is the way memory is distributed between the ESNs.
%Nevertheless, the correlation coefficients of the signals shown in Figures \ref{fig:esn_1_output} and \ref{fig:esn_2_output} do not appear to match with the observed similarity of the signals in Figures \ref{fig:esn_1_output} and \ref{fig:esn_2_output}. A possible explanation for this mismatch might be that a lot of the trained signals show delay, which may distort the correlation coefficient and make the signals appear uncorrelated. 

%Judging from the similarity in shape of the observed signals as well as the close correlation of the signal clusters to each other, in combination with the computed NMSE losses we concluded that backpropagation is suitable to decompose tasks on multi-reservoir systems.

The final evaluation is the transferability of the learned signals. We again trained 10 new randomly generated multi-reservoir systems with identical architecture and ESN hyperparameters, now using the intermediate signals of the best performing architecture trained with backpropagation (Net 0) as desired signals for the intermediate reservoir stages. This gives us an average NMSE of 0.056 on the test set which is only slightly worse than the performance of the backpropagated system from which the desired signals have been derived. The rather low standard deviation of $0.008$ on the test loss also indicates that the derived desired signals work for a wider range of ESNs with random weights. Here, future work will be directed at finding appropriate ways to regularise the backpropagation learning, in order to avoid overfitting to the specific reservoir instances that are used. This should help to close the remaining transferability gap.

\section{Conclusion}
In this paper we have proposed a new training method that can be used to decompose machine learning tasks onto a given multi-reservoir network. It uses backpropagation to derive desired signals for training the individual reservoirs in the network. We established that a specialized engineered multi-ESN architecture can outperform a general-purpose single reservoir system for the chosen task. While jointly training the whole multi-ESN system using backpropagation does not necessarily outperform the classic single ESN system due to varying convergence of the training algorithm, it appears to be suitable to automatically decompose the NARMA-10 task into subtasks. In some of the cases, the learned decomposition approximates the quasi-optimal decomposition derived by hand, where this happens more frequently for ESNs closer to the output. Finally, using the derived intermediate signals as desired signals for a classic training approach delivers similar performance as obtained through backprop. 

The experiments in this paper were designed to offer a first proof-of-concept of the approach. Although the results are not yet competitive, they show that the approach is promising. Future work will be directed at the two critical points of the approach: the ability to efficiently backpropagate through multiple layers of reservoirs and minimising the amount of overfitting to the individual reservoir instances used during backpropagation. Previous work \cite{hermans2015trainable} indicates that the first is in principle feasible, especially in view of the fact that they have far less nonlinearities than the ESNs used in this study. For the second, many options are still unexplored.

This leads us to conclude that our proposed method will likely be useable to derive targets in multi-reservoir integrated photonic reservoir systems, which we believe to be the most technologically feasible way to use them on more challenging tasks than what is currently possible. We will transfer this work to photonic reservoir computing, using the recently released PhotonTorch framework \cite{laporte2019highly} for the backpropagation and validate it on actual devices.

\balance



\bibliographystyle{IEEEtran}
\bibliography{IEEEabrv,nlinv}

\end{document}